\documentclass[10pt]{article}
\usepackage{titling}
\usepackage{fullpage} 
\usepackage[size=small,format=plain,labelfont=up,textfont=up]{caption}
\usepackage[letterpaper, margin=1.0in]{geometry}
\usepackage{bm,dsfont,amssymb,amsmath}
\usepackage{subfigure}
\usepackage{titlesec}
\usepackage{color}
\usepackage{url}
\usepackage{enumitem}
\usepackage{multirow}
\usepackage{multicol}
\usepackage[normalem]{ulem}
\usepackage[labelfont=bf,singlelinecheck=false]{caption}
\usepackage{hyperref}
\usepackage{graphicx}

\title{\vspace{-2cm}m-TSNE: A Framework for Visualizing High-Dimensional                      Multivariate Time Series}

\vspace{-30pt}
\author{
Minh Nguyen$^1$, Sanjay Purushotham, PhD$^1$, Hien To$^1$, Cyrus Shahabi , PhD$^1$\\
$^1$University of Southern California, Los Angeles, CA, USA\\
\vspace{-0.25in}
}
\date{}

\begin{document}
\pagenumbering{gobble} 
\maketitle
\vspace{-0.5in}
\section*{Abstract}
\textit{
Multivariate time series (MTS) have become increasingly common in healthcare domains where human vital signs and laboratory results are collected for predictive diagnosis. Recently, there have been increasing efforts to visualize healthcare MTS data based on star charts or parallel coordinates. However, such techniques might not be ideal for visualizing a large MTS dataset, since it is difficult to obtain insights or interpretations due to the inherent high dimensionality of MTS. In this paper, we propose "m-TSNE”: a simple and novel framework to visualize high-dimensional MTS data by projecting them into a low-dimensional (2-D or 3-D) space while capturing the underlying data properties. 
Our framework is easy to use and provides interpretable insights for healthcare professionals to understand MTS data. We evaluate our visualization framework on two real-world datasets and demonstrate that the results of our m-TSNE show patterns that are easy to understand while the other methods’ visualization may have limitations in interpretability. 
}

\section{Introduction}
\label{sec:introduction}
Big data analytics in healthcare is emerging as a large amount of health and medical data are being generated every day. Recent development of different types of health sensors and e-health platforms has opened up great opportunities for collecting, monitoring and analyzing patients' health conditions from multiple data sources~\cite{raghupathi2014big}. Performing analytics and extracting insights on healthcare data is challenging due to the large volume, high dimensionality, heterogeneity, and dynamic nature of the healthcare data. To address these challenges, many studies are being conducted in several fields such as machine learning, data mining, statistics, health informatics, etc. Data visualization is one such field that provides tools for visual interpretations of the underlying data patterns and trends, and helps in further data analysis.

Several techniques \cite{van1999cluster,aris2005representing, li2012visualizing, shi2012rankexplorer} have been developed to visualize and analyze time series data. Since these techniques usually analyze univariate time series (UTS) data by handling only one data variable at a time (e.g., monitoring respiratory rate or heart rate over time to detect anomaly), they may not fully capture the inherent correlations of the multivariate time series (MTS) data \cite{yang2004pca}. Therefore, we believe that MTS data corresponding to multiple variables should be treated as a whole, rather than being broken into individual UTS as they can provide greater insight into data representing patients' conditions. 
However, multivariate data is notoriously hard to represent because of the difficulty of mentally picturing data in more than three dimensions~\cite{heer2010tour}.

\begin{figure}[ht]
	\begin{minipage}[b]{0.50\linewidth}
		\centering
	\includegraphics[width=1\textwidth]{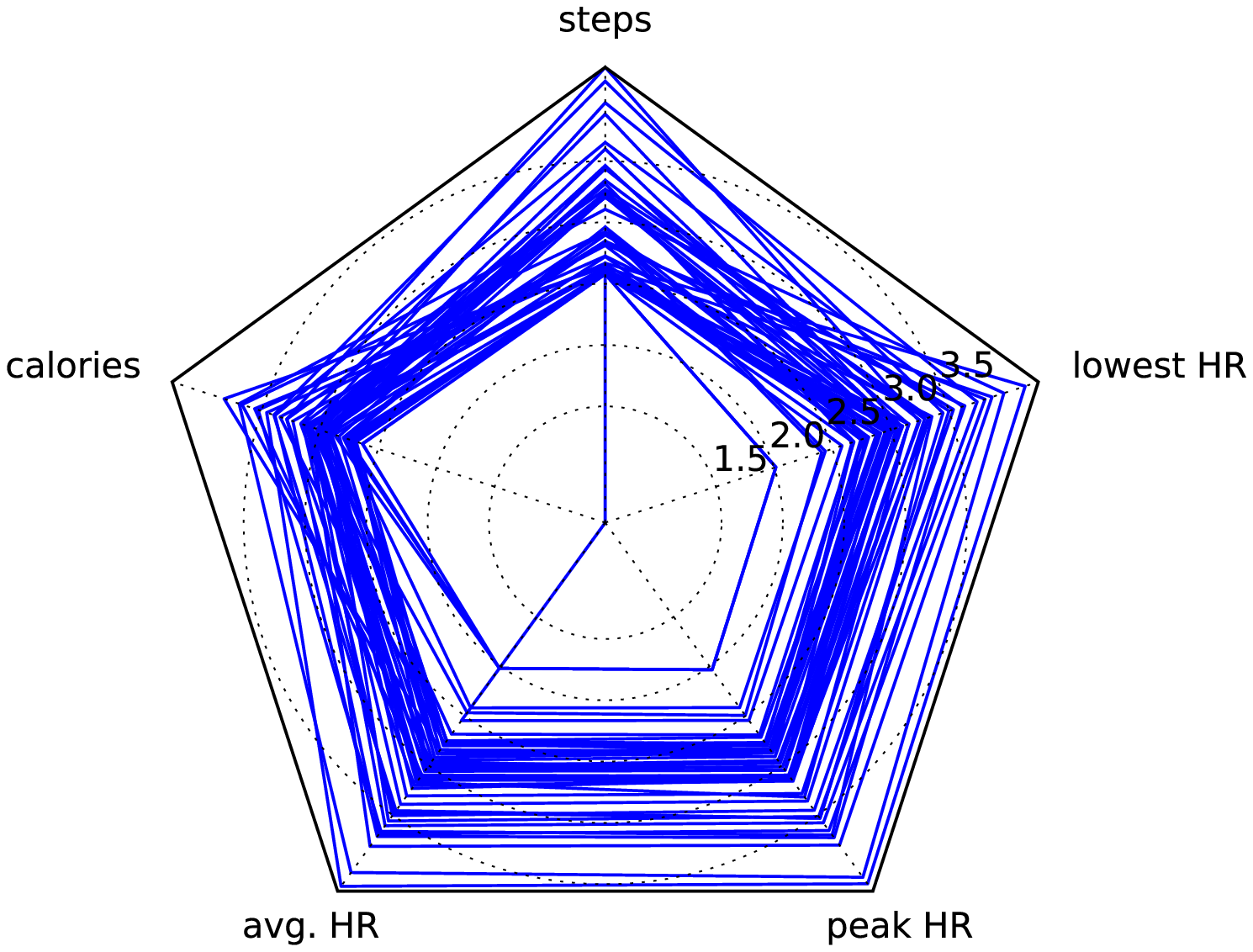}
		\caption{Star chart visualization \cite{chambers1983graphical} of MTS human monitoring data of a subject in ATOM-HP dataset (Section~\ref{sec:datasets}). Data has 5 variables: step counts, calories, lowest heart rate, average heart rate, and peak heart rate. 
        Each polygon is a multivariate data point at a time instance.}
		\label{fig:star_plot}
	\end{minipage}
		\hspace{3pt}
	\begin{minipage}[b]{0.48\linewidth}
		\centering
	\includegraphics[width=1\textwidth]{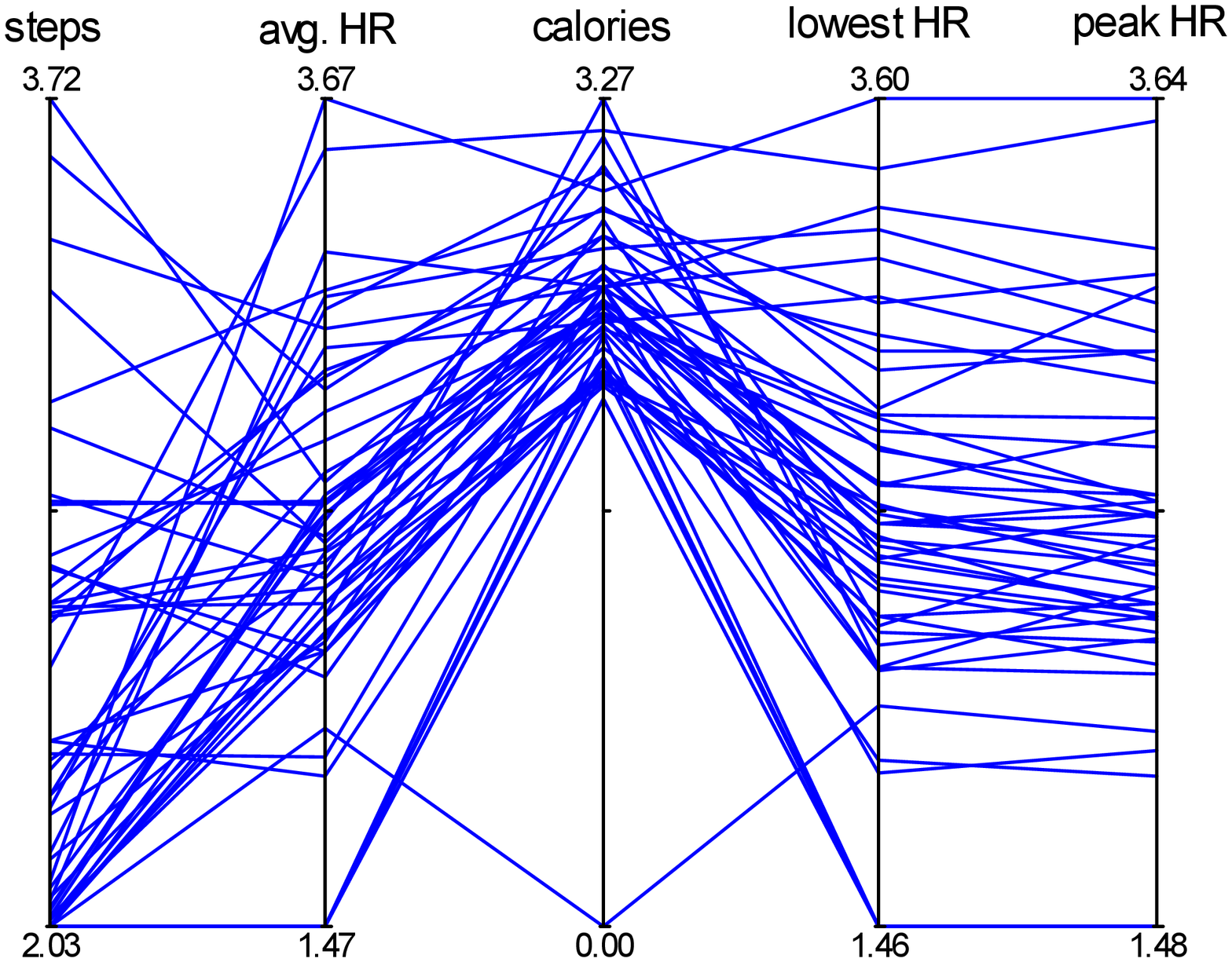}
		\caption{Parallel coordinates visualization \cite{inselberg1985plane} of the same data shown in Figure \ref{fig:star_plot}. 
        Each polyline that connects the parallel axes represents a multivariate data point at a time instance. The number of polyline in the figure is the number of time instance in the MTS.}
		\label{fig:parallel_plot}
	\end{minipage}
\end{figure}

Previous visualization efforts \cite{chambers1983graphical, inselberg1985plane,lehmann2008visualizing, west2013visualization} on multivariate data focused on displaying multiple dimensions of the data in 2-D plot, and left the interpretations to human observer. Figures \ref{fig:star_plot} and \ref{fig:parallel_plot} show the star chart and parallel coordinates visualizations that have been previously proposed for healthcare data visualization \cite{lehmann2008visualizing,west2013visualization}. These techniques might not be ideal for real-world applications which have high-dimensional MTS data points since they tend to result in complex visualization plots with less interpretability. Researchers have explored dimensionality reduction techniques \cite{jolliffe2002principal,hinton2002stochastic,maaten2008visualizing} to handle high-dimensional data points by projecting them into low-dimensional space. These approaches  have been applied successfully in computer vision applications where data consists of images. However, such techniques, in general, cannot be applied directly on medical MTS datasets because each data source may have different properties, e.g. time dependency structure is present in vital signs time series. Hence, there is a need for developing new techniques to visualize high-dimensional MTS data. 



In this paper, we propose m-TSNE (Multivariate Time Series t-Distributed Stochastic Neighbor Embedding): a framework for visualizing MTS data in low-dimensional space that is capable of providing insights and interpretations of the high-dimensional MTS datasets. m-TSNE first calculates the similarity between each MTS data points in high-dimensional space, based on Extended Frobenius norm (EROS) \cite{yang2004pca} which is a similarity metric for MTS data. Then, it computes the low-dimensional (2-D or 3-D) projection of the MTS data points using a gradient descent method by preserving the similarity relation between pairs of high-dimensional points. We conduct visualization experiments on two healthcare datasets: ATOM-HP dataset (Section \ref{sec:datasets}), a dataset collected to study impact of chemotherapy on patient's activity; and Electroencephalogram (EEG) dataset~\cite{zhang1995event}, a dataset collected to study whether there is a genetic predisposition to alcoholism.
We evaluate and the compare m-TSNE's visualizations with other competing approaches by conducting an user study. We show that m-TSNE's visualizations on the ATOM-HP dataset can extract patient activity level patterns and outliers (during chemotherapy cycle) which helps oncologists to study their treatment's effects on patient's fatigue. On the EEG dataset, we show that control and alcoholic subjects can be easily identified (separated) in our low-dimensional visualization which is not possible in the original high-dimensional space.      


The organization of the paper is as follows: We discuss the related work in the field of data visualization in Section \ref{sec:related}. Section \ref{sec:ourapproach} describes our m-TSNE visualization framework. Section~\ref{sec:experiments} reports the empirical evaluation using the two aforementioned datasets and Section \ref{sec:discussions} concludes with discussion and future work.

\section{Related Work}
\label{sec:related}

\begin{figure}
\centering
\includegraphics[height=4cm]{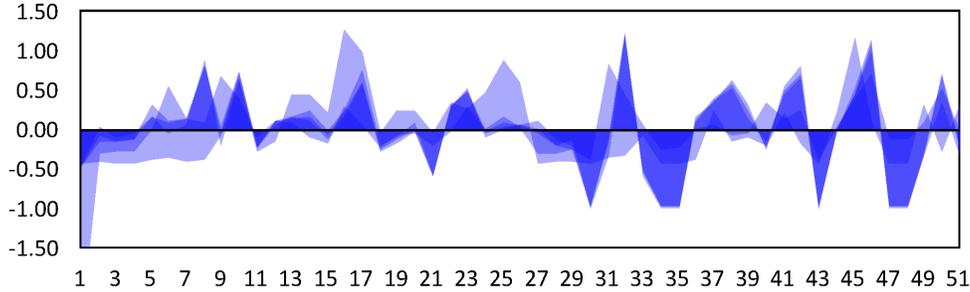}
\caption{MTSA visualization~\cite{lehmann2008visualizing} of the same raw data as Figure \ref{fig:star_plot}. x-axis represents the time instance (the date number), and y-axis represents the variables values.}
\label{fig:mtsa}
\vspace{-12pt}
\end{figure}

Due to the pervasiveness of time series data and its wide range of applications in healthcare monitoring, stock market analysis, traffic analysis, etc., understanding time-series data through visualization has attracted lots of attention, among which there have been efforts focusing on visualizing MTS data \cite{chambers1983graphical,inselberg1985plane,lehmann2008visualizing,borland2014multivariate}. These techniques visualize multivariate data by simply displaying individual variable data independently on one shared-space display \cite{javed2010graphical} without treating multiple variables as a whole. Nevertheless, shared-space techniques may not be efficient to visualize high-dimensional MTS dataset since the resulting shared-space visualization may show a large amount of overlapping and clutter between different variables' time series, which is difficult to interpret. For instance, as shown in Figure~\ref{fig:mtsa}, in \cite{lehmann2008visualizing}, the authors proposed a technique named Multivariate Time Series Amalgam (MTSA), to jointly visualize multiple variables of MTS on a single display.
Nevertheless, MTSA may not be suitable when the MTS dataset has a large number of measurements or has high dimensionality. The authors in \cite{borland2014multivariate} proposed a visualization method for multivariate data based on star chart \cite{chambers1983graphical} which represents each variable on an axis, with the axes arranged around a circle. An example of this technique is depicted in Figure \ref{fig:star_plot}. Another popular technique for visualizing MTS is parallel coordinates \cite{inselberg1985plane} (shown in Figure \ref{fig:parallel_plot}). These methods have limitations in interpretability when the MTS has a large number of variables or a large number of time instances per variable. 
Unlike these studies, our proposed m-TSNE technique deals with multiple variables of MTS as a whole by applying dimensionality reduction techniques to project MTS data points to low-dimensional latent space and it also preserves the distances between the MTS in the original space.

\section{MTS Visualization Framework}
\label{sec:ourapproach}

\begin{figure}
\centering
\includegraphics[width=\textwidth]{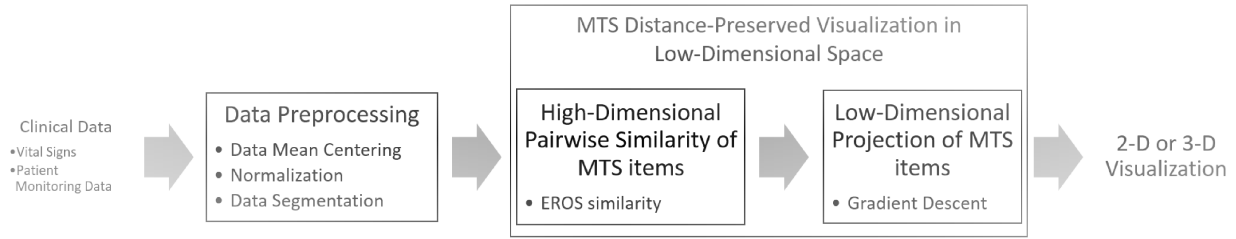}
\caption{Pipeline of m-TSNE framework for MTS Visualization}
\label{fig:pipeline}
\vspace{-10pt}
\end{figure}

Figure \ref{fig:pipeline} shows the pipeline of our proposed m-TSNE approach  for MTS Visualization. First, the MTS data is processed by mean-centering and normalization; then it is segmented into multiple MTS items \cite{jolliffe2002principal} (discussed in Section \ref{sec:preprocess}). EROS  \cite{yang2004pca} is calculated to find the high-dimensional pairwise similarity between the MTS items (see Section \ref{sec:similarity}). For visualization, the MTS items is projected to low-dimensional space with the guarantee that the high-dimensional pairwise similarity relation is preserved (explained in Section \ref{sec:projection}). That is, if two MTS items are neighbors (or far apart) in high-dimensional space, their low-dimensional projection points should also be close (or far apart) to each other. We use gradient descent method similar to \cite{maaten2008visualizing} for dimensionality reduction and projection.

\subsection{Data Preprocessing}
\label{sec:preprocess}
Data preprocessing consists of data mean-centering, normalization, and segmentation. Mean-centering by subtracting the mean value of a variable, and normalization by dividing the variable with its standard deviation helps the MTS items to be in the same scale \cite{jolliffe2002principal} which is needed for the further MTS similarity calculation step. The MTS data is also segmented into MTS items corresponding to the time instance (eg. an hour, a day). The trend of the data over the time should be visualized as the patterns of all MTS items.

\subsection{MTS Pairwise High-dimensional Similarity}
\label{sec:similarity}
One of the widely used techniques for high-dimensional data similarity calculation is t-Distributed Stochastic Neighbor Embedding (t-SNE)~\cite{hinton2002stochastic}. In order to calculate the high-dimensional data points similarity, t-SNE \cite{hinton2002stochastic}  computes the Euclidean distance between two data points, then converts the distance into conditional probabilities that represent similarities. The details of the techniques can be found in \cite{nguyen2016vahc}.

As our study concerns MTS, the similarity metric should be suitable for MTS data type and should also take into account the correlation between MTS variables. The Euclidean distance similarity metric is shown to be not suitable for time series data as acceleration and deceleration along the time axis is suboptimal for distance matching \cite{maimon2005data}. As a result, Dynamic Time Warping (DTW) is generally used to overcome the limitations of Euclidean distance metric \cite{muller2007dynamic}. However, both of these distance only work for UTS data matching, as they do not consider the correlation between MTS variables. Therefore, the t-SNE technique may not perform well on MTS datasets if Euclidean distance or DTW is used as a similarity metric. 

We propose to use EROS similarity metric \cite{yang2004pca} to overcome the shortcomings for above similarity metrics. EROS similarity metric is a technique based on the Principal Component Analysis method in which an MTS item is treated as a whole, i.e. it is not broken in multiple UTS, to preserve the correlation between variables. Let us denote MTS data by $ X $, where $X = \{ x_1, x_2,...,x_k \}$ and $x_i$ corresponds to a multivariate data point (high-dimensional) at time instance $ i $. Each $x_i$ is termed as an MTS item and it is represented as an $m$ x $n$ matrix, where $m$ is the number of observations (eg. patients), and $n$ is the number of variables. Each variable can be a vital sign variable such as heart rate, respiration rate, etc. or an activity monitoring variable such as step count, active hour, etc. Given $2$ MTS items $x_i$ and $x_j$, EROS first computes the eigenvectors and eigenvalues of each item. Thereafter, it measures the cosine similarities of the corresponding eigenvectors of $x_i$ and $x_j$. Finally, EROS similarity is the weighted sum of all the cosine similarities of the eigenvectors. The weight is calculated as the aggregated value based on all the eigenvalues of the MTS items in the dataset. 
The EROS similarity metric is described in Equation \ref{eq:eros}: 
\begin{equation} \label{eq:eros}
EROS(x_i, x_j, w) = \sum_{l=1}^{n} w_l |<v_{il}, v_{jl}>|
\end{equation}
Where, $EROS(x_i, x_j, w)$ is the similarity of MTS item $x_i$ and MTS item $x_j$, $v_i = [v_{i1}, ··· , v_{in}]$ and $v_j = [v_{j1}, ··· , v_{jn}]$ are the two sets of eigenvectors of $x_i$ and $x_j$ respectively and $w_l$ is the aggregated weight computed based on the eigenvalue corresponding to the $l^{th}$ eigenvector in the weight vector $w$.
The computed pairwise similarities of MTS items are used for projecting each MTS item into low-dimensional (2-D or 3-D) space. Section \ref{sec:projection} describes the details of how to project MTS data points to lower dimensional space using a gradient descent method.

\subsection{MTS Low-dimensional Projection}
\label{sec:projection}
Principal Component Analysis (PCA) \cite{jolliffe2002principal,krzanowski1979between} is a popular approach to map high-dimensional data into low-dimensional space. PCA is a linear mapping which focuses on preserving the low-dimensional projection of dissimilar data points far apart. However, PCA is shown not to be suitable for non-linear manifolds where preserving the low-dimensional projection of similar data points close to each other is important \cite{hinton2002stochastic,maaten2008visualizing}.
To overcome this drawbacks of PCA, we propose to perform an optimization method: gradient descent (similar to the one used in t-SNE~\cite{maaten2008visualizing}) for m-TSNE where we minimize the mismatch between high-dimensional and low-dimensional spaces. The gradient descent method works by minimizing a cost function over all data points. The details of the cost function and performing gradient descent is explained in our technical report \cite{nguyen2016vahc}.
In the following section, we will discuss the empirical results of our m-TSNE approach, and compare it to PCA, Euclidean-based t-SNE, and DTW-based t-SNE approaches for MTS visualization.

\section{Experiments}
\label{sec:experiments}
\subsection{Datasets}
\label{sec:datasets}
We evaluate m-TSNE visualization using two healthcare datasets: ATOM-HP\footnote{ATOM-HP: Analytical Technologies to Objectively Measure Human Performance} dataset which is a coarse-grained time series dataset collected from our on-going study, and EEG dataset - a fine-grained time series dataset from UCI machine learning repository \cite{Lichman:2013}.

\textbf{ATOM-HP dataset} : This dataset is collected to study how to quantify the activity levels of cancer patients undergoing chemotherapy treatment to complement the current clinical assessment. The patients suffer from treatment induced fatigue that affects their daily activities, which is usually measured by physicians using Eastern Cooperative Oncology Group (ECOG) scores \cite{oken1982toxicity}. However, this score tends to suffer from subjective bias. Therefore, a more robust objective measurement is needed to evaluate patients' activity / performance status, which is the goal of our study. In ATOM-HP, each patient carries a wearable sensor during their chemotherapy treatment cycle. The chemotherapy cycle consists of two chemotherapy visits. The date range between the two chemotherapy visits varies from two to three weeks. There are eight patients in our dataset (more patients are being enrolled in this on-going study). The daily data of each patient is an MTS which has five variables: number of steps, total calories, average heart rate, peak heart rate, and lowest heart rate. Collected data is sampled every hour and for each patient, data is collected for at least 50 consecutive days.

\textbf{EEG dataset} \cite{zhang1995event} : The dataset is collected to examine if EEG correlates genetic predisposition to alcoholism. There are three versions of the dataset. In our work, we use the large dataset version that has 10 control subjects, and 10 alcoholic subjects. Each subject performs 30 trials which can be classified as three trial types: exposure to a single stimulus, exposure to two matching stimulus, and exposure to two non-matching stimulus \cite{snodgrass1980standardized}. In total there are 600 (=20x30) trials. The data of one trial is an MTS of 64 variables corresponding to 64 electrodes placed on the subject's scalps. The data sample rate is 256Hz.

\subsection{Experimental Setup}
\label{sec:setup}
To evaluate our framework, we compare m-TSNE to the visualizations of PCA, Euclidean-based t-SNE, and DTW-based t-SNE methods. In the ATOM-HP dataset, given an MTS data of one subject, we are interested in visualization of MTS to show the trends and outliers in the subject's daily activity performance during the chemotherapy treatment. The subject's MTS data is represented as an $m$ x $5$ matrix with $m$ is the total hours of monitoring, and $5$ is the number of variables (Section \ref{sec:datasets}). For instance, the size of the MTS matrix data of the subject in Figure \ref{fig:band_3d} is $1224$ x $5$. As the study considers monitoring daily performance, the MTS is segmented into multiple MTS items of $24$ x $5$ matrices which represent the 24 hours of the data after mean-centering, and normalization. In Figure \ref{fig:band_3d}, we have a total of 51 ($=1224/24$) data points corresponding to 51 MTS items data matrices collected during 51 days when the subject was involved in the study. 
For m-TSNE, we calculate the pairwise similarity using EROS similarity metric (Equation \ref{eq:eros}). The pairwise similarities are put through the gradient descent to compute the MTS items' projection in low-dimensional (2-D or 3-D) space (Figure \ref{fig:band_3d}, Figure \ref{fig:band_2d}). For the EEG dataset, each subject's trial is represented as an  MTS matrix of $256$ x $64$ where 256 is the number of observations and 64 is the number of EEG variables. The pairwise EROS similarities of 600 MTS items corresponding to 600 trials is calculated. The similarities are then used to compute the projection of 600 MTS as Figure \ref{fig:eeg_3d}.




\textbf{PCA}: The MTS is preprocessed, and segmented into MTS items as above (Section \ref{sec:setup}). After preprocessing, the daily aggregated multivariate data points are computed as sum of number of steps, sum of calories,  average heart rate, peak heart rate, and lowest heart rate over 24 hours. All aggregated data of MTS items are put together as a matrix of $m'$ x $5$ where $5$ is the number of variables, and $m'$ is the number of data points with each row representing an aggregated multivariate data point. 
PCA is used for low-dimensional projection and  visualization as shown in Figure \ref{fig:band_2d_pca}.

\textbf{Euclidean-based t-SNE}: After data preprocessing, a pairwise Euclidean distance for each data points pair is calculated. Based on the Euclidean distance, a pairwise probability similarity matrix of each pair of data points is computed as in \cite{nguyen2016vahc}. 

\textbf{DTW-based t-SNE}: Since Euclidean distance does not work well for time series data, we can measure MTS similarity using DTW distance. This approach is similar to Euclidean-based t-SNE, but uses the pairwise Dynamic Time Warping distance instead of Euclidean distance.

\textbf{m-TSNE}: m-TSNE calculates EROS pairwise similarity matrix, and computes MTS projection using the gradient descent method as described in Section \ref{sec:ourapproach}. 

%

We implemented above four methods in Python 2.7.11 and we will release our code on GitHub.


\subsection{Visualization Results}
\label{sec:results}
\begin{figure}[ht]
	\begin{minipage}[b]{0.49\linewidth}
		\centering
	\includegraphics[width=1\textwidth]{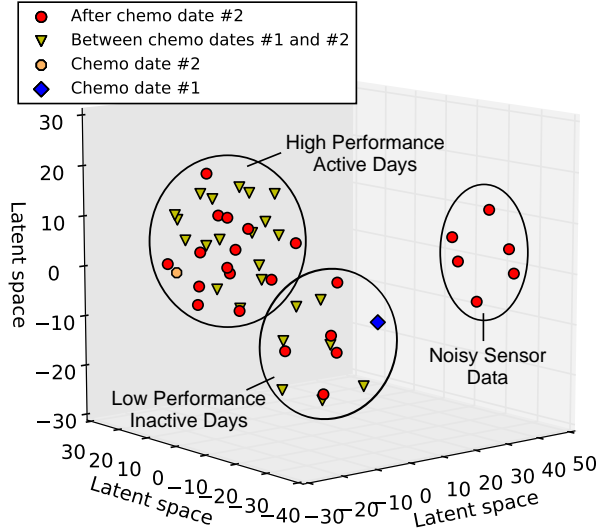}
		\caption{m-TSNE visualization of ATOM-HP dataset.}
		\label{fig:band_3d}
                \vspace{-10pt}
	\end{minipage}
		\hspace{3pt}
	\begin{minipage}[b]{0.49\linewidth}
		\centering
	\includegraphics[width=1\textwidth]{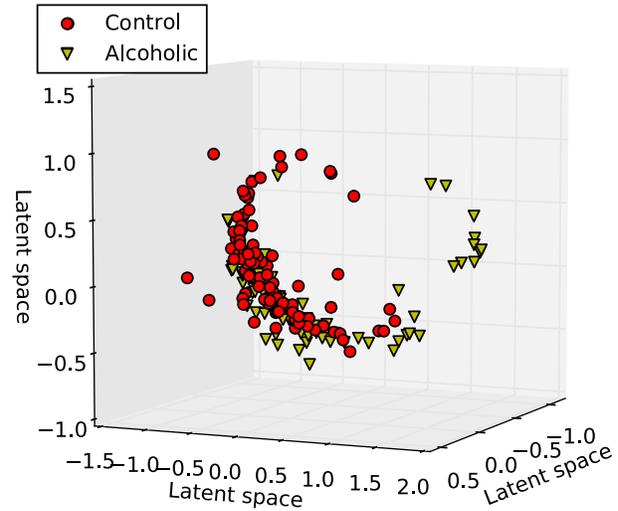}
		\caption{m-TSNE visualization of EEG dataset.}
		\label{fig:eeg_3d.eps}
        \vspace{-10pt}
	\end{minipage}
\end{figure}
\begin{figure}[ht]
	\begin{minipage}[b]{0.49\linewidth}
		\centering
	\includegraphics[width=1\textwidth]{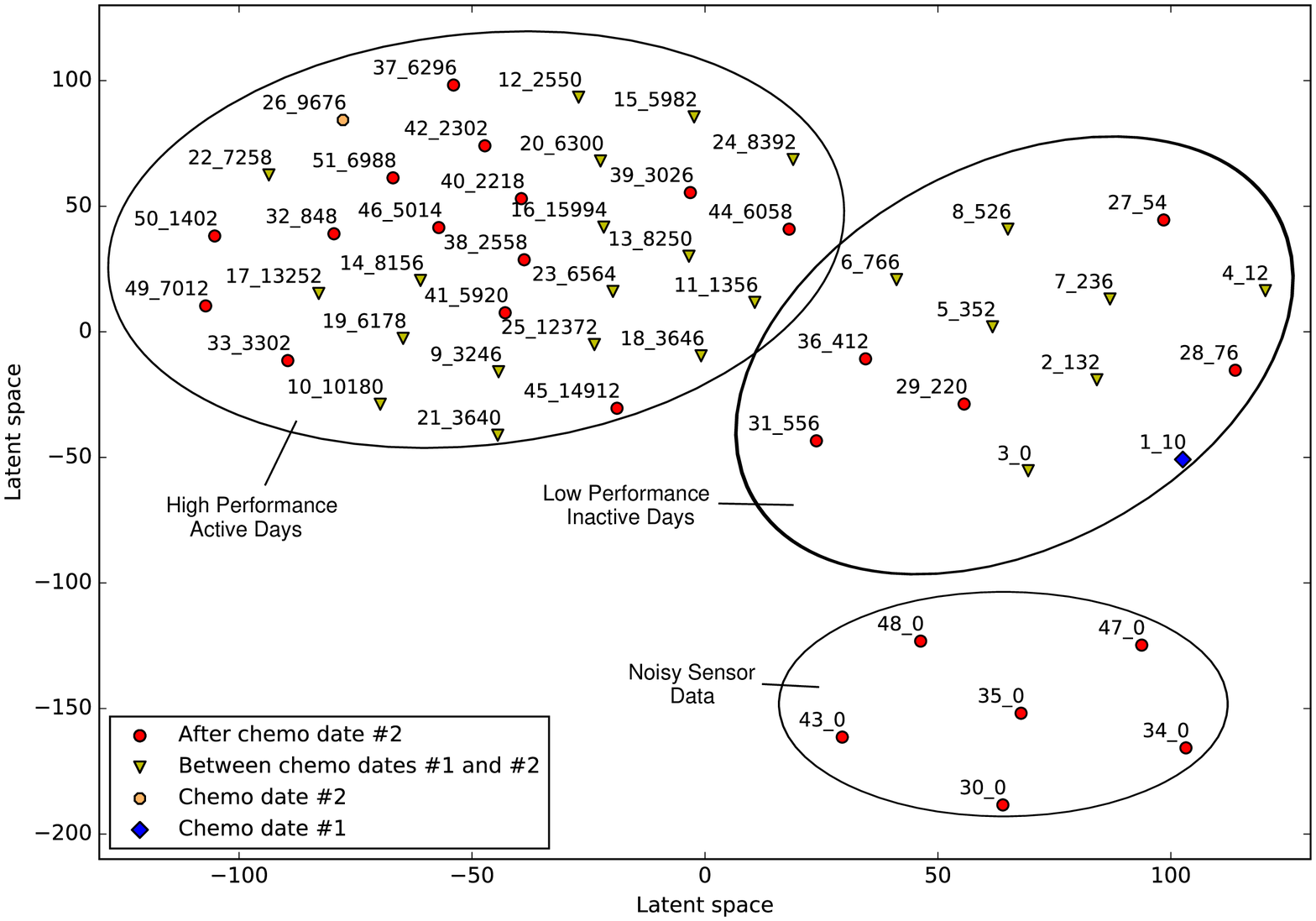}
\caption{m-TSNE 2-D visualization of ATOM-HP dataset with annotations.}
		\label{fig:band_2d}
	\end{minipage}
		\hspace{3pt}
	\begin{minipage}[b]{0.49\linewidth}
		\centering
	\includegraphics[width=1\textwidth]{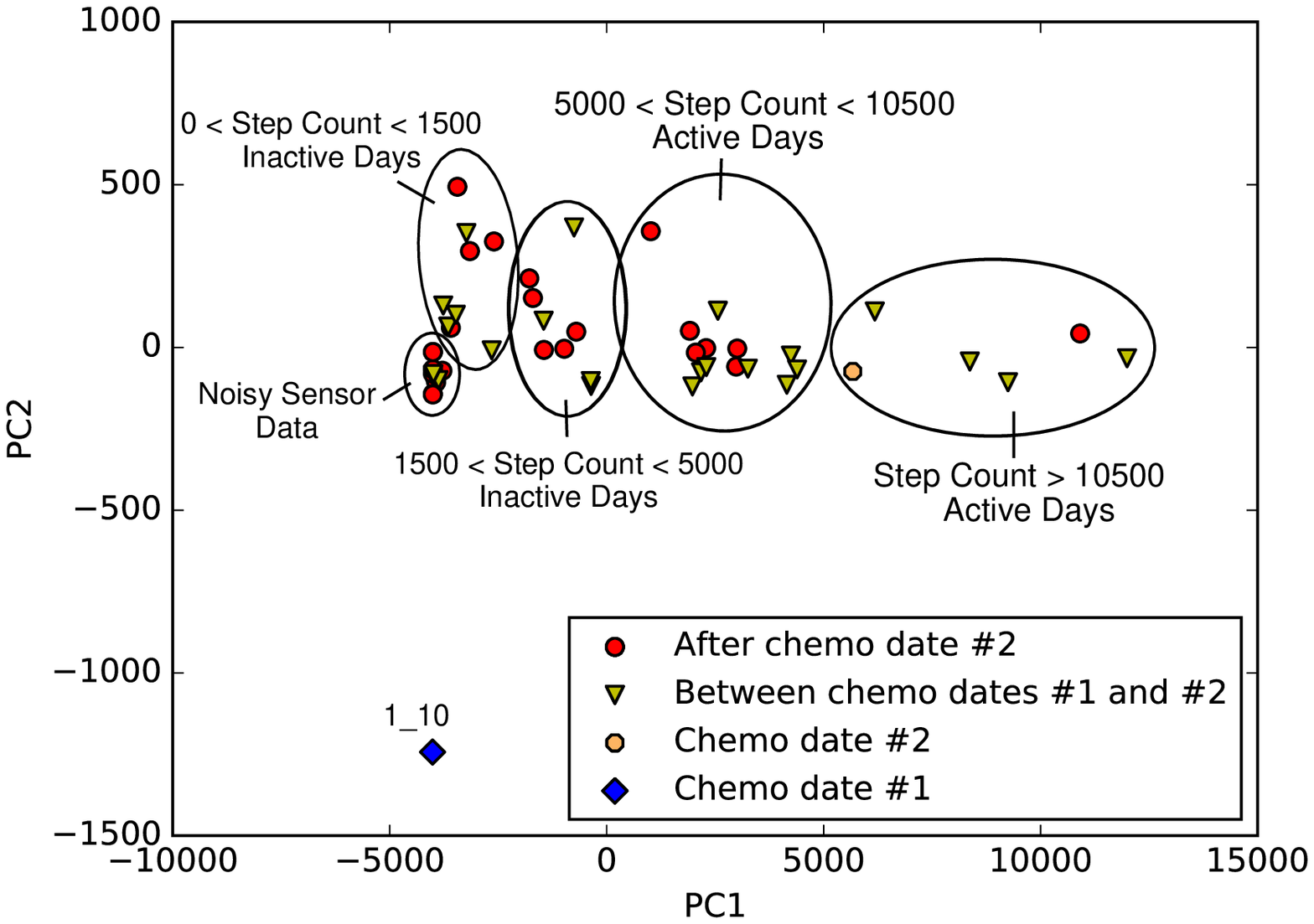}
\caption{PCA 2-D visualization of ATOM-HP dataset.\\}
\label{fig:band_2d_pca}
	\end{minipage}
\end{figure}

Figure \ref{fig:band_3d} shows that the visualization of the MTS items of a subject undergoing treatment using m-TSNE. In this figure, there are $51$ points corresponding to 51 days of chemotherapy treatment cycle. Each data point is labeled by the date order whether it is before or after a chemotherapy date in the study cycle. As can be seen, the figure shows that the $51$ points form $3$  clusters. To understand the three clusters, a 2-D display with annotation for each data point is provided in Figure \ref{fig:band_2d}. The annotation are in a format $[\textbf{date\ number}]$\textunderscore$[\textbf{step\ counts}]$ (step counts variable is chosen as it is easier to interpret human performance using this variable as compared to other variables). The figure was also shown to a health professional. Based on the health professionals' inputs, and the variables values, in Figure \ref{fig:band_3d}, these clusters can be interpreted as: (1) cluster of high performance / active days (the left-most cluster with a high number of step counts for each data point (more than 800)), (2) cluster of low performance / inactive days (the lower cluster with a low number of step counts for each data point), (3) cluster of noisy data (outliers) from the sensors (the right-most cluster) (noisy data appears to have abnormal values due to sensors). This is one of the insights obtained from our visualization approach that can help the healthcare professionals to understand the effect of the treatment using data collected from wearable sensors. Moreover, the figure also shows that the subject has low activities for a few days immediately following the two chemotherapy dates. This insight indicates that the subject may suffer from fatigue due to the chemotherapy session. We believe that these insights are quite helpful for oncologists to study their patient's activity performance, and also these might help them in designing better objective measures to quantify human performance. 

\begin{figure}[ht]
	\begin{minipage}[b]{0.49\linewidth}
		\centering
	\includegraphics[width=1\textwidth]{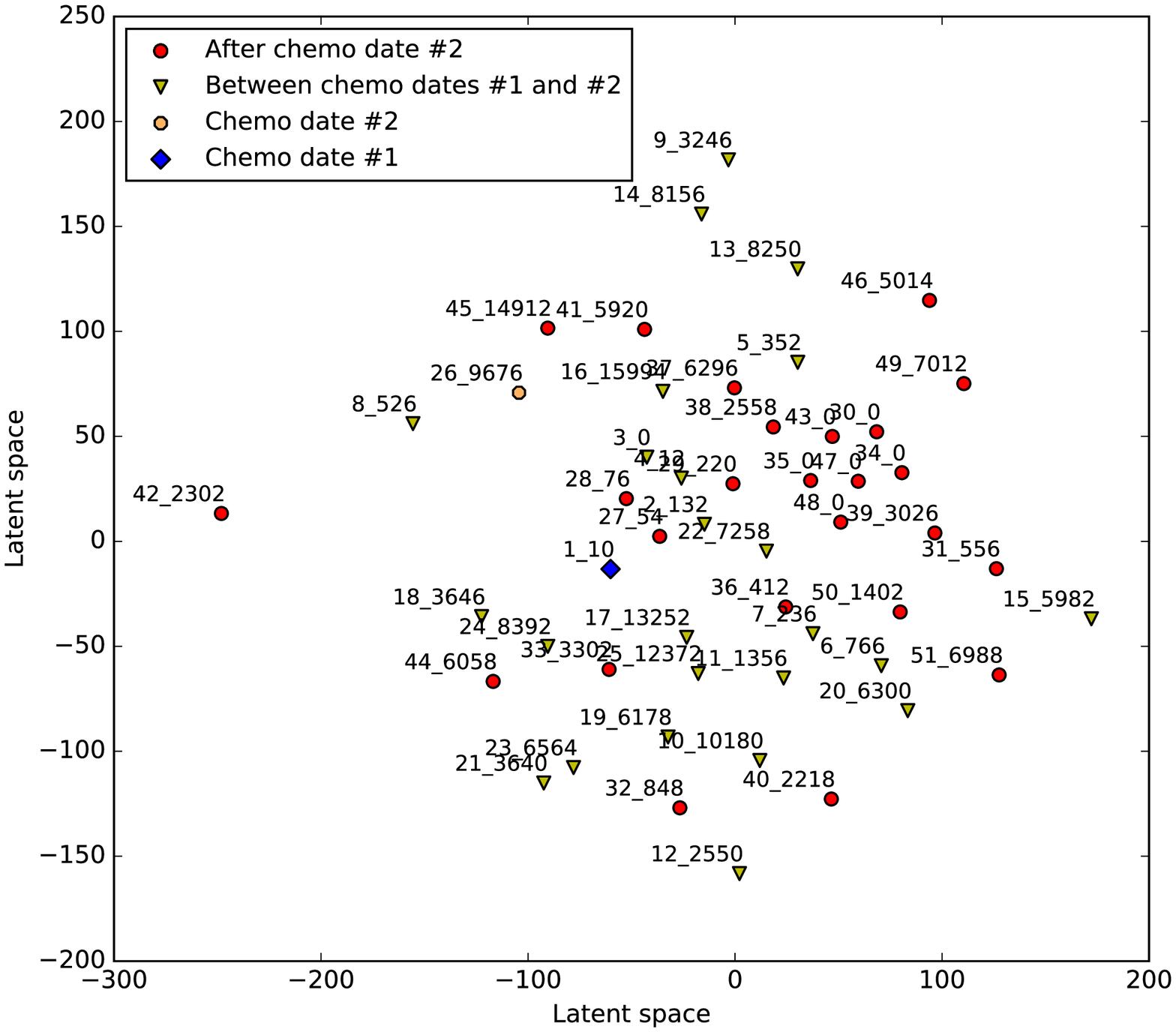}
\caption{Euclidean-based t-SNE 2-D visualization of ATOM-HP dataset with annotations.}
		\label{fig:band_2d_ed}
	\end{minipage}
		\hspace{3pt}
	\begin{minipage}[b]{0.49\linewidth}
		\centering
	\includegraphics[width=1\textwidth]{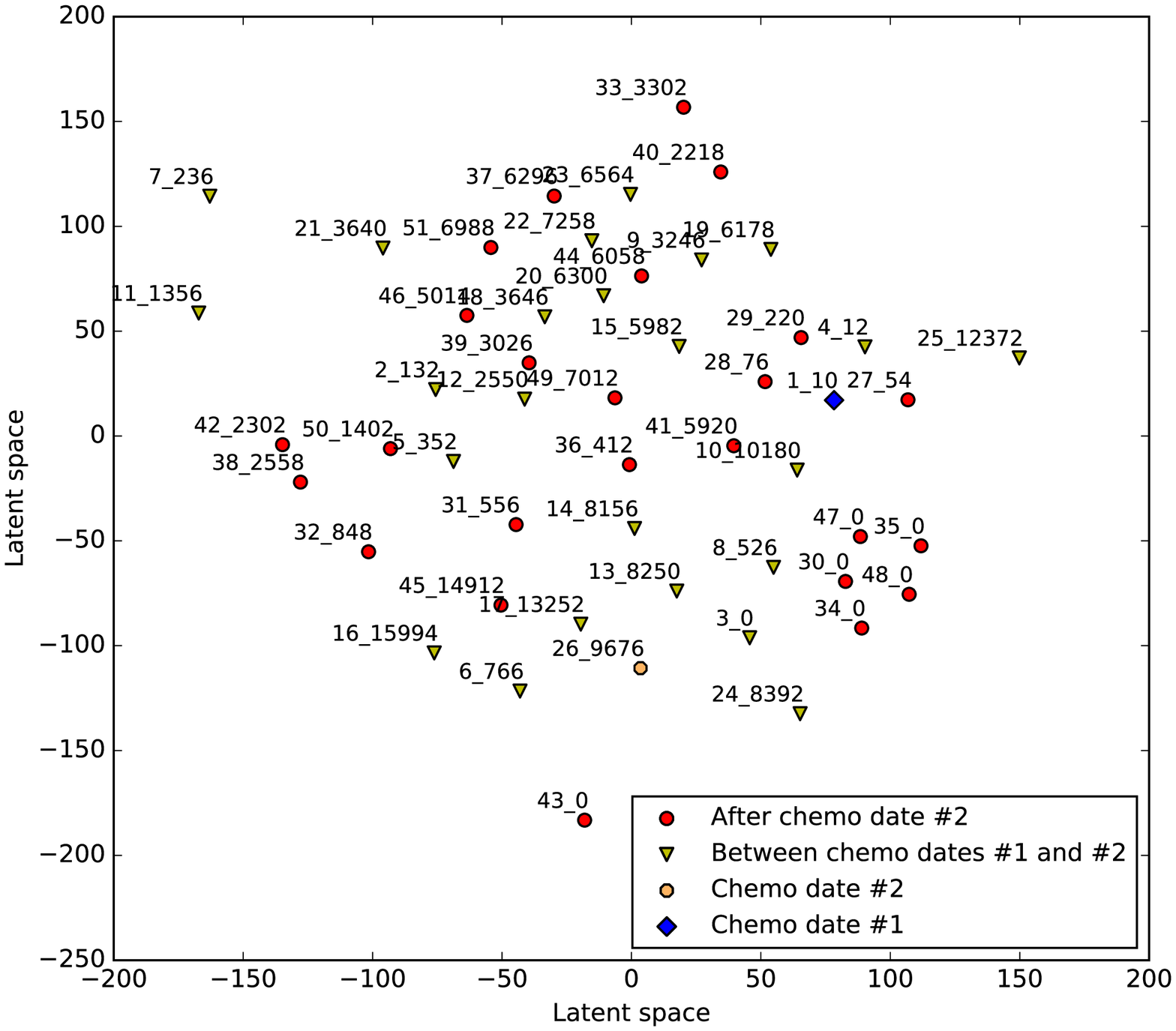}
\caption{DTW-based t-SNE 2-D visualization of ATOM-HP dataset with annotations.}
\label{fig:band_2d_dtw}
	\end{minipage}
\end{figure}

For comparison, we provide the visualizations of the same subject using PCA, Euclidean-based t-SNE and DTW-based t-SNE methods in Figure \ref{fig:band_2d_pca}, Figure \ref{fig:band_2d_ed}, and Figure \ref{fig:band_2d_dtw}, respectively. Note, the star chart, parallel coordinates chart, and MTSA of the same subject's data are also shown in Figure \ref{fig:star_plot}, Figure \ref{fig:parallel_plot}, and Figure \ref{fig:mtsa} respectively. It is clear that these visualizations do not provide clear insights about the patient's activities / performance status.
Euclidean-based, and DTW-based t-SNE figures (Figure \ref{fig:band_2d_ed}, Figure \ref{fig:band_2d_dtw}) show that the noisy sensor data points are close to each other. However, they do not give clear distinct clusters like m-TSNE. Based on these figures, it may be difficult to interpret and understand the data insights as these approaches show overlapping clusters which appear as a cloud of data points. PCA (Figure \ref{fig:band_2d_pca}) might provide some insights of the subject performance along the axis of highest principle component, however as shown in the figure it does not form distinct clusters for outliers or for different levels of activity.





Figure \ref{fig:eeg_3d} provides m-TSNE visualization for the EEG dataset. Each data point is one subject's MTS data performing a trial, and is labeled based on the subject category: \emph{control} and \emph{alcoholic}. The figure clearly depicts a manifold in which the points representing the control group lie inside, and are covered by alcoholic group. It also shows that all the outliers in our visualization belong to alcoholic group, providing interpretable insights which are not extracted by PCA or the other MTS visualization techniques.


\textbf{User Study:} We evaluated the interpretability of three techniques: PCA, DTW-based t-SNE and m-TSNE by conducting a controlled user study with 6 non-healthcare professionals. We removed the names of the techniques, labels and color codes from the visualization results to avoid bias in  users' interpretability. Each user was shown the visualizations of these three techniques on all subjects in ATOM-HP dataset and instructed to assign a score if they could find (interpretable) clusters, trends or outliers. For each technique, an user assigned a score (1 - the lowest score, 2 or 3 - the best score) in such a way that better techniques received higher scores. We aggregated the scores over all users and report out findings here: m-TSNE obtained the highest score of 2.48, PCA obtained a score of 1.92, and DTW-based t-SNE obtained the lowest score of 1.6. 
An oncologist was also included in our user study to verify our visualizations and insights. He agreed that the clusters found by our approach is very useful to study the patients' fatigue during their treatment cycle. This user study shows that m-TSNE could provide interpretability and insights when compared to the other competing methods.



\section{Conclusion and Future Work}
\label{sec:discussions} 
In this paper, we proposed m-TSNE: a framework to visualize high-dimensional MTS data. m-TSNE uses EROS to compute similarity between MTS data points and projects them to low-dimensional space for visualization. Empirical evaluation on two healthcare datasets showed that our approach provides interpretable insights via visualization while the other visualization methods which use PCA, Euclidean-based t-SNE, and DTW-based t-SNE, are more difficult to interpret. These insights could help healthcare professionals to evaluate their patients' performance. For future work, we plan to extend our work by building a tool for showing the visualization of the MTS dynamically.


\paragraph{Acknowledgments:} This research has been funded in part by the National Cancer Institute (award number P30CA014089), National Institutes of Health, Department of Health and Human Services, under Contract No. HHSN261201500003B, and the USC Integrated Media Systems Center. Any opinions, findings, and conclusions or recommendations expressed in this material are those of the authors and do not necessarily reflect the views of any of the sponsors.

\bibliographystyle{IEEEtran}
\begin{footnotesize}
\bibliography{references}
\end{footnotesize}

\end{document}